\documentclass[11pt]{article}

\usepackage[table,xcdraw]{xcolor}
\usepackage[preprint]{acl}

\usepackage{times}
\usepackage{latexsym}
\usepackage{todonotes}
\usepackage{amsmath}
\usepackage{amssymb}
\usepackage{amsfonts}
\usepackage{booktabs}
\usepackage{tcolorbox}
\tcbuselibrary{breakable}
\usepackage{microtype}
\usepackage{enumitem}
\usepackage{subcaption}
\usepackage[T1]{fontenc}
\usepackage{multirow}
\usepackage{algorithm}
\usepackage{algpseudocode}
\usepackage{amsmath} %
\algrenewcommand\algorithmicindent{1.0em}

\usepackage[utf8]{inputenc}

\usepackage{microtype}

\usepackage{inconsolata}

\usepackage{graphicx}

\title{ProCeedRL: Process Critic with Exploratory Demonstration Reinforcement Learning for LLM Agentic Reasoning}

\author{
 \textbf{Jingyue Gao\textsuperscript{1,3}},
 \textbf{Yanjiang Guo\textsuperscript{1,3}},
 \textbf{Xiaoshuai Chen\textsuperscript{2}},
 \textbf{Jianyu Chen\textsuperscript{1,3}}
\\
 \textsuperscript{1}Institute for Interdisciplinary Information Sciences, Tsinghua University, \\
 \textsuperscript{2}Independent Researcher,
 \textsuperscript{3}Shanghai Qizhi Institute
\\
 \small{
   \href{mailto:gaojy23@mails.tsinghua.edu.cn}{gaojy23@mails.tsinghua.edu.cn}
 }
}

\begin{document}
\maketitle
\begin{abstract}

Reinforcement Learning (RL) significantly enhances the reasoning abilities of large language models (LLMs), yet applying it to multi-turn agentic tasks remains challenging due to the long-horizon nature of interactions and the stochasticity of environmental feedback.
We identify a structural failure mode in agentic exploration: suboptimal actions elicit noisy observations into misleading contexts, which further weaken subsequent decision-making, making recovery increasingly difficult.
This cumulative feedback loop of errors renders standard exploration strategies ineffective and susceptible to the model's reasoning and the environment's randomness.
To mitigate this issue, we propose \textbf{ProCeedRL}: \textbf{Proc}ess \textbf{C}ritic with \textbf{E}xplorativ\textbf{e} \textbf{D}emonstration RL, shifting exploration from passive selection to active intervention.
ProCeedRL employs a process-level critic to monitor interactions in real time, incorporating reflection-based demonstrations to guide agents in stopping the accumulation of errors.
We find that this approach significantly exceeds the model's saturated exploration performance, demonstrating substantial exploratory benefits.
By learning from exploratory demonstrations and on-policy samples, ProCeedRL significantly improves exploration efficiency and achieves superior performance on complex deep search and embodied tasks.

\end{abstract}

\section{Introduction}

\begin{figure}[htbp]
    \centering
    \includegraphics[width=\linewidth]{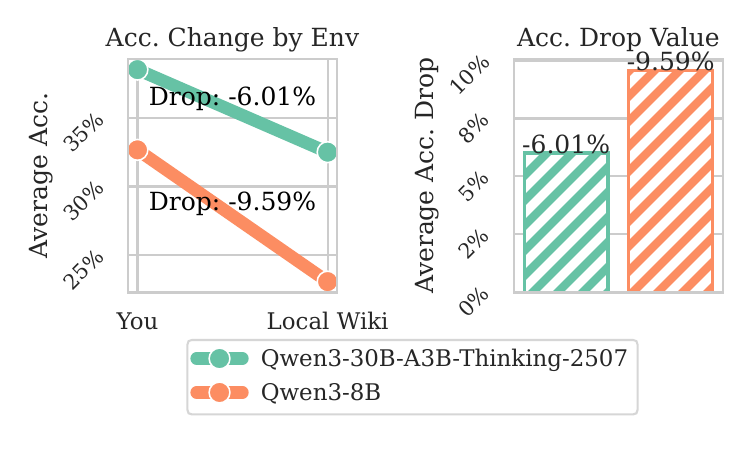}
    \vspace{-25pt}
    \caption{
    In multi-turn agentic tasks, we find that noisy environmental feedback in the context can degrade a model’s reasoning ability (left), with weaker models being affected more severely (right).
    These two phenomena make recovery in multi-turn interactions extremely difficult, as accumulated contextual noise rapidly compounds and quickly weakens the model.
    }
    \vspace{-10pt}
    \label{fig:illustrative_exp}
\end{figure}

Large Language Models (LLMs) like DeepSeek-R1~\citep{deepseekai2025deepseekr1incentivizingreasoningcapability} exhibit exceptional reasoning capabilities, driven by Reinforcement Learning with Verifiable Rewards (RLVR)~\citep{shao2024deepseekmathpushinglimitsmathematical,deepseekai2025deepseekr1incentivizingreasoningcapability}.
This paradigm, which optimizes LLMs with reward based on the correctness of the final output, has proven effective for single-turn problems such as mathematics~\citep{openai2024openaio1card,kimiteam2025kimik15scalingreinforcement}, and agentic reasoning tasks like tool-integrated reasoning recently~\citep{ragen,feng2025retoolreinforcementlearningstrategic,song2025r1searcherincentivizingsearchcapability,jin2025search,singh2025agenticreasoningtoolintegration}.

However, multi-turn agentic reasoning introduces additional challenges for standard RLVR exploration due to its long-horizon, stochastic nature.
Agents must continuously interact with stochastic environments, in which the history of feedback is incorporated into the context for subsequent reasoning.
Such a workflow creates a high-stakes dependency between agents and environments.
Suboptimal actions, such as a vague search query due to limited agent capability, elicit irrelevant or misleading feedback from the environment, which is then added to the context.
The accumulated context noise, in turn, influences and degrades subsequent reasoning and action quality.
This results in a vicious circle in agentic exploration.
Any suboptimal action or environmental stochasticity may trigger and progressively reinforce this loop of accumulated error, making recovery increasingly difficult.
Therefore, the upper bound of standard RLVR exploration achieved through repeated sampling is limited by the agents and environments.

To analyze how noise in context affects model capability, we compare different models on search-augmented question answering under varying levels of environmental noise.
Fig.~\ref{fig:illustrative_exp} shows that both models are sensitive to environment feedback and experience a performance drop in the noisier environment, suggesting the impact of observation quality on agent reasoning.
Notably, we find that the weaker model exhibits a greater decline in performance.
The amplified degradation indicates that limited reasoning capabilities exacerbate the impact of noisy observations.
The adverse effects of suboptimal actions accumulate through the feedback loop between the agent and the environment, making recovery difficult.
This validates the susceptibility of vanilla agentic exploration to agents and environments, underscoring its importance for agentic reasoning.

To mitigate this problem, we propose \textbf{ProCeedRL}: \textbf{Pro}cess \textbf{C}ritic with \textbf{E}xplorativ\textbf{e} \textbf{D}emonstra\-tion \textbf{RL}.%
Rather than waiting to penalize negative trajectories after collecting repeated samples, ProCeedRL employs a process-level critic to detect suboptimal steps in real time.
By rewinding the agent to retake the step with a refined demonstration before proceeding, ProCeedRL prunes invalid actions and breaks the vicious circle of accumulating error between actions and environments.
It addresses the vicious circle in exploration and enables the model to learn beyond its inherent exploration limit.
Furthermore, learning from these refined demonstrations at negative steps embeds this knowledge into the model's inherent capabilities, thereby eliminating the need for additional pipelines at test time.

Several related works also share a similar idea of providing process supervision during agentic reasoning from different viewpoints.
\citet{wang2025stepsearch,deng2025atomsearcherenhancingagenticdeep} introduce process-level rewards based on extracted documents or fine-grained actions, which requires costly data labeling and reward design.
Another line of concurrent work~\citep{li2025reseekselfcorrectingframeworksearch,fu2025re} proposes different formulations of self-correcting search pipelines, including self-proposed sub-goals or an agent-invoking judging pipeline.
However, models are trained to use these frameworks at test time, whereas our method internalizes the critique information into the model's inherent capabilities.

\begin{figure*}[htbp]
    \centering
    \includegraphics[width=\linewidth]{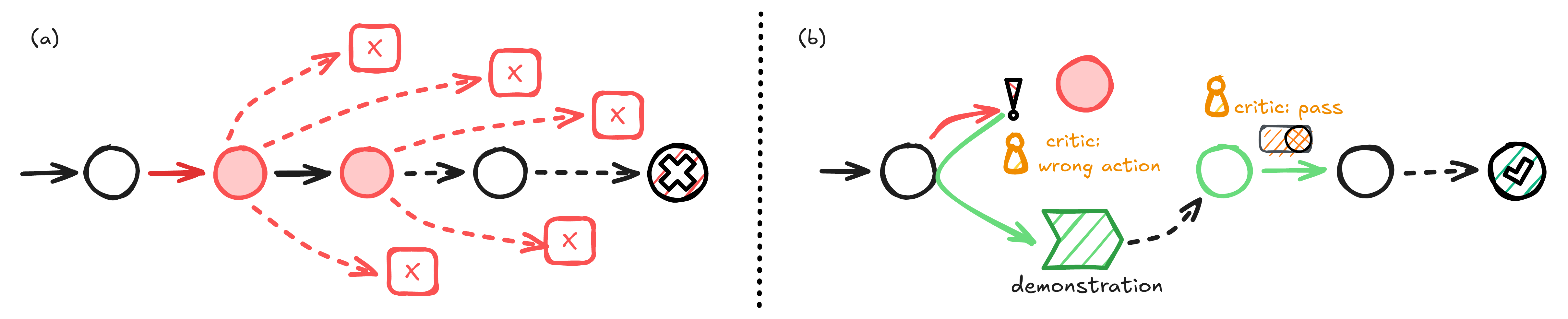}
    \caption{Comparison between standard independently repeated sampling (a) and ProCeedRL rollouts (b).
    \textbf{Left}: In vanilla exploration, the model's suboptimal action may result in irrelevant or misleading observations, which hinder all subsequent reasoning and the exploration of correct samples.
    \textbf{Right}: In ProCeedRL, a critic actively monitors the planning process.
    When an adverse action is detected due to faulty reasoning or low-quality returned observations, a refined demonstration replaces it to guide the agent out of the vicious circle in (a) and mitigate the exploration issue.
    }
    \label{fig:overview_banner}
    \vspace{-1em}
\end{figure*}

We conduct comprehensive experiments and analyses in agentic reasoning to validate the effectiveness of our method.
Compared to baselines, ProCeedRL significantly improves exploration efficiency and surpasses the model's upper bound in standard RLVR.
In summary, our contributions are as follows:
\begin{itemize}[itemsep=-2pt, topsep=5pt, leftmargin=1.2em]
  \item Identifies the cumulative interaction between suboptimal actions and environmental noise, which mutually amplify into a vicious cycle that renders standard exploration inefficient;
  \item Proposes ProCeedRL, a framework that shifts from passive selection to active intervention by using process critics to rewind and refine errors;
  \item Show the effectiveness of ProCeedRL in improving exploration and surpassing the reasoning limit on deep search and embodied tasks.
\end{itemize}

\section{Related Works}

\paragraph{Reinforcement Learning in LLMs}
Reinforcement Learning with Verifiable Rewards (RLVR) excels in enhancing the reasoning abilities of LLMs on complex tasks like mathematical reasoning and code generation~\citep{deepseekai2025deepseekr1incentivizingreasoningcapability, shao2024deepseekmathpushinglimitsmathematical,qwq32b}.
Recent work~\citep{ragen,feng2025group,feng2025retoolreinforcementlearningstrategic,chen2025research} extends RLVR with outcome rewards to improve agentic reasoning, such as deep search and tool use. %
However, these methods rely on independently repeated generation, which limits exploration efficiency and reasoning upper bound constrained by the base model~\citep{yue2025doesreinforcementlearningreally}.
ProCeedRL improves exploration by actively leveraging the critic and refining actions at the step level, thereby surpassing an agent's inherent limits. 

\paragraph{Process Supervision for Reasoning}
Process supervision yields dense rewards and better credit assignment for LLM reasoning.
\citet{lightman2024lets,zhang2025lessons,wang2023math} propose and discuss how to develop process reward models.
Another line of work~\citep{dong2025arpo, Kazemnejad2024:VinePPO,gao2025marge,chen2025research,wang2025stepsearch,deng2025atomsearcherenhancingagenticdeep} uses carefully designed rewards or pipelines to provide process supervision for reasoning without PRMs.
Among them, some concurrent works~\citep{li2025reseekselfcorrectingframeworksearch, fu2025re} design test-time reflection pipelines with specially designed rewards to encourage it.
In contrast, ProCeedRL incorporates refined demonstrations during training that serve as implicit guidance, avoiding the cumbersome design of rewards or additional overhead at test time.

\section{Method}

\begin{figure*}[htbp]
    \centering
    \includegraphics[width=\linewidth]{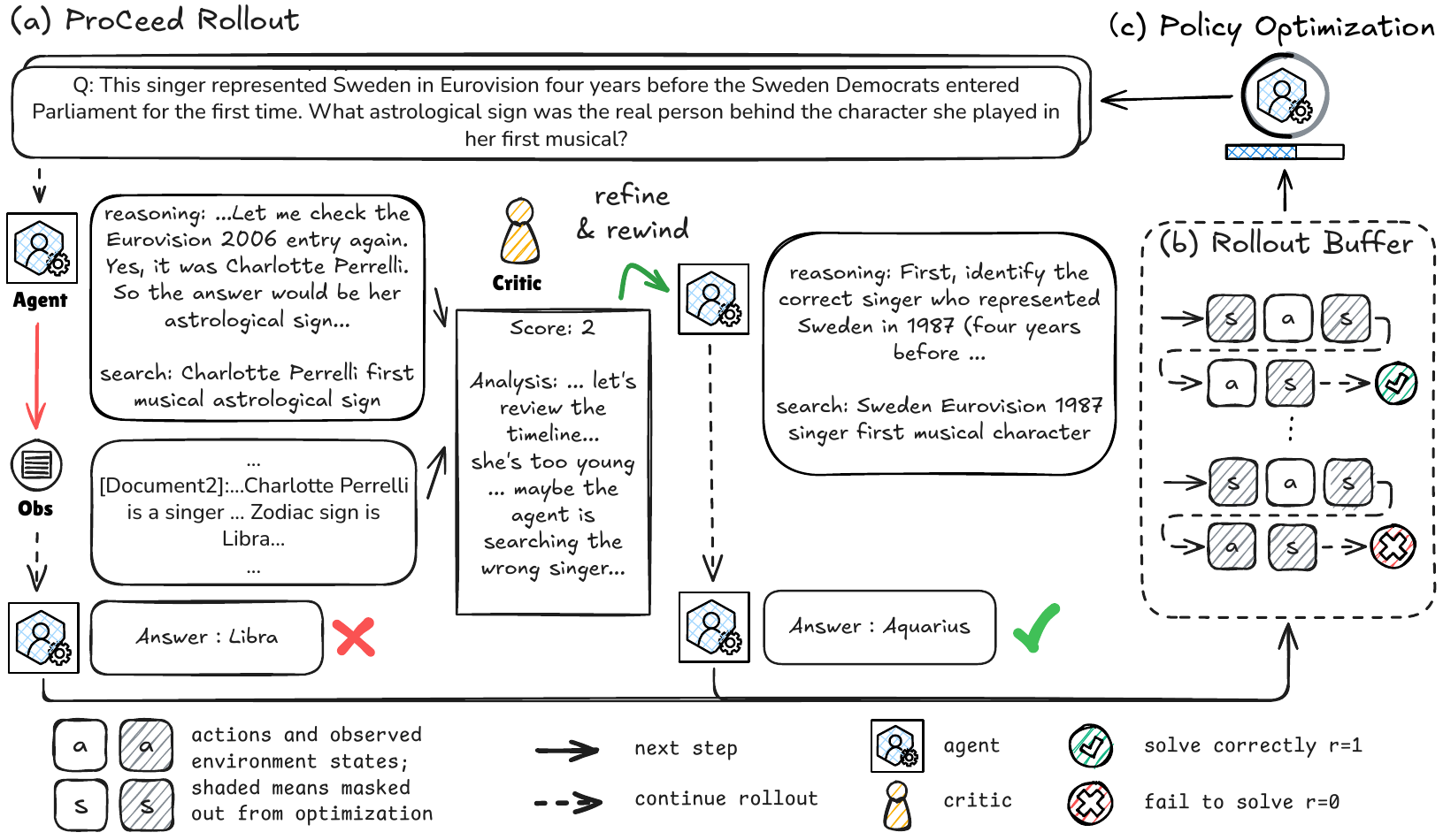}
    \caption{The overall workflow of our method with an example.
    \textbf{(a)} When the agent makes a suboptimal action, the observation reinforces this path and derails subsequent model reasoning, resulting in low exploration efficiency (Sec.~\ref{sec:method_problem}).
    We employ a process-level critic to identify flawed steps, in which the agent refines its actions and reruns the adverse actions, thereby breaking the vicious circle and improving exploration and reasoning limits (Sec.~\ref{sec:method_framework}).
    \textbf{(b)} After collecting trajectories with ProCeed rollout, we incorporate them with directly generated samples to form meaningful groups for subsequent policy optimization in \textbf{(c)}, as detailedly described in Sec.~\ref{sec:method_theory}.
    }
    \label{fig:main}
    \vspace{-0.5em}
\end{figure*}

In this section, we introduce our method, beginning with the formulation of the agentic reasoning problem in Sec.~\ref{sec:method_problem_formulation}.
We then analyze a key problem in agentic reasoning in Sec.~\ref{sec:method_problem} and present our solution with the ProCeedRL framework in Sec.~\ref{sec:method_framework}.
Sec.~\ref{sec:method_theory} outlines the overall training pipeline of our method, additionally with an overview and a case example in Fig.~\ref{fig:main} and the algorithm in Alg.~\ref{alg:main}.

\subsection{Preliminaries}
\label{sec:method_problem_formulation}

\paragraph{Problem Settings}
Given a multi-turn reasoning task $s_0$ in environments, we define a trajectory as $\tau=(s_0, a_0, r_0, \cdots)$, where both $s_t$ and $a_t$ are textual strings representing the agent's observed state and action, respectively.
At each timestep $t$, the LLM agent $\pi_\theta$, receives the observed state $s_t$ and produces action $a_t$ according to policy $\pi_\theta(\cdot|\tau_t)$.
The environment yields the next state $s_{t+1}$ and a reward $r_t\in\mathbb{R}$ based on $a_t$.
We adopt an outcome-verified reward setting: the environment assigns a terminal reward of $r(\tau)=1$ upon successful completion (e.g., a correct answer or successful task completion), and $r_t=0$ for all intermediate steps.

\paragraph{LLM RL Algorithms}
We optimize policy $\pi_\theta$ to maximize the expected reward $\mathbb{E}_{\tau\sim\pi{\theta}(\tau)} r(\tau)$ using RL algorithms like GRPO~\citep{shao2024deepseekmathpushinglimitsmathematical} and DAPO~\citep{yu2025dapoopensourcellmreinforcement}.
While these methods leverage group-based relative advantage for stability, they rely on vanilla repeated sampling for exploration.
Building on advances in RLVR, our work addresses the exploration challenge in self-generated rollouts by incorporating process-level critic and refinement signals into the rollout.

\subsection{Vicious Circle in Agentic Reasoning}
\label{sec:method_problem}

We identify a critical problem in agentic reasoning: a downward spiral between suboptimal actions and misleading environment observations.
Since agent context is cumulative of actions and observations, it creates a dependency between the agent's actions and the observations it receives.
Any errors in actions, due to incorrect reasoning or hallucination, introduce misleading feedback into the context.
The ``poisoned'' context, in the meantime, derails subsequent reasoning and leads to suboptimal actions.
It reinforces a positive feedback loop that progressively degrades final performance, which is inevitable given the agent's reasoning ability and the environment's randomness.

To validate this hypothesis, we conduct an illustrative experiment (Fig.~\ref{fig:illustrative_exp}) comparing two models with distinct reasoning capabilities, \texttt{Qwen3-8B} and \texttt{Qwen3-30B-A3B-Thinking-2507}, on search-augmented QA.
We simulate varying environmental noise by using different search engines: one is a commercial search engine, You\footnote{https://you.com/}, while the noisier one is a local dense retriever based on the Wikipedia corpus\footnote{https://huggingface.co/datasets/PeterJinGo/wiki-18-corpus}.
We evaluate the average accuracy of both models on Bamboogle, Frames, and WebwalkerQA across the two environments.

The findings first validate the premise that a noisy context degrades the model's reasoning, as both models suffer performance drops with the local retriever across three benchmarks.
In particular, the weaker reasoner (\texttt{Qwen3-8B}) exhibits a more pronounced decrease.
This susceptibility confirms that environmental noise amplifies action errors, corroborating the existence of a vicious circle between suboptimal actions and corrupted reasoning contexts.
To mitigate this, we employ a critic to monitor suboptimal actions and refine them in real time to avoid adverse effects, thereby improving exploration efficiency and outcomes.

\subsection{ProCeed Rollout}
\label{sec:method_framework}
In this section, we present the core design of the ProCeedRL rollout phase, illustrated in Fig.~\ref{fig:overview_banner}. We detail how our method identifies and rectifies errors in real time, thereby preventing error propagation.

\paragraph{Process Level Critic}
To provide step-granular supervision and intercept potential reasoning errors, we employ a critic $\phi$ to evaluate the quality of each intermediate step.
Formally, at each timestep $t$, we ask the critic to output an integer scalar score $l_t$ and a textual critique $c_t$, based on both history $\tau$ and the latest interaction $a_t, s_{t+1}$: $l_t, c_t = \phi(\tau_t,a_t,s_{t+1})$.

By incorporating the subsequent observation $s_{t+1}$, the critic can ground its evaluation in the action's actual effect rather than just its plausibility.
Specifically, for irreversible environments, such as the real world, we evaluate $a_t$ directly without $s_{t+1}$.
An action $a_t$ is deemed adverse if its score falls below a specific threshold, $l_t \leq l_\text{th}$.
In practice, we calibrate $l_\text{th}$ dynamically across tasks according to the critic and agent's behavior, as discussed in the ablation in Sec.~\ref{sec:ablation_study}.

\paragraph{Refined Exploratory Demonstration}
Upon detecting an adverse step, we trigger an intervention to break the ``vicious cycle'' of misleading observations.
Instead of allowing the agent to proceed with a suboptimal action $a_t$, we perform an on-the-fly rectification.
We utilize a refining policy $\mu$ to generate a refined action $a^\prime_t$, conditioned on the critic's specific feedback: $a_t^\prime = \mu(\tau_{t-1}, a_t,l_t,c_t)$.

We then replace the original action $a_t$ with the refined demonstration $a_t^\prime$.
They help the agent avoid suboptimal actions and break the vicious cycle of potentially misleading contexts, thereby exhibiting exploratory benefits.
Crucially, our framework is model-agnostic: while $\phi$ and $\mu$ can be instantiated with strong external models to maximize supervision quality, we empirically find that using the policy model $\pi_\theta$ itself is also highly effective.
This enables ProCeedRL to operate as a scalable, self-contained pipeline that is independent of external knowledge.
Prompts for both the critic and refinement modules are detailed in Appendix~\ref{appendix:prompt}.

\subsection{ProCeedRL training}
\label{sec:method_theory}

\paragraph{Training ProCeedRL}
The training process for ProCeedRL, depicted in Fig.~\ref{fig:main}, extends the standard group-based RLVR framework.
We augment each data group in the replay buffer with critic scores and refined demonstrations, while also including trajectories sampled directly from the current policy.
These direct samples serve as negative and on-policy references, which are crucial for group-based RLVR algorithms.
The model is subsequently trained on this buffer using an optimization algorithm such as SFT or DAPO.
Notably, we mask the demonstration steps in trajectories that fail during optimization since these demonstrations are off-policy relative to the model.
Reducing their probability may cause instability and confusion during optimization.

\paragraph{Policy Optimization}
The demonstration steps collected through refinement do not align with the current policy distribution, resulting in a distributional shift between the training data and the target policy.
Directly incorporating misaligned demonstrations $d_i$ with on-policy samples $o_i$ into the RL objective negatively affects performance~\citep{yan2025learningreasonoffpolicyguidance,zhang2025onpolicyrlmeetsoffpolicy}.
We adopt the method chord-$\phi$~\citep{zhang2025onpolicyrlmeetsoffpolicy}, which applies a coefficient to weight the demonstration loss, down-weighting both highly probable and unlikely demonstrations.
Specifically, for a rollout buffer $\mathcal{B}=\{o_i\}\cup\{d_j\}$, the optimization object in our method when using DAPO is:
\begin{equation}
\scalebox{0.9}{$
\begin{aligned}
     \mathbb{E}_{x\sim \mathcal{B}, \{o_i\}\sim \pi_{\theta_\text{old}}, \{d_{j}\}\sim \mu}
\Bigg[\frac{1}{|G|}\sum_{i=1}^{|G|}\sum_{t=1}^{|d_i|} \sigma(d_{j,t}) \hat{A}_{i,t}\min&\\ \Big( 
\text{is}_{i,t}(\theta),
\ \text{clip} \Big( \text{is}_{i,t}(\theta), 1 - {\varepsilon_{\text{low}}}, 1 + {\varepsilon_{\text{high}}} \Big) \Big) \Bigg]&,
\end{aligned}
$}
\label{eq:optimization}
\end{equation}
where $A_i = \frac{r_i-\text{mean}(r_i)}{\text{std}(r_i)}$ is the relative advantage for response $i$, $\text{is}_{i,t}(\theta) = \frac{\pi_{\theta}(o_{i,t} \mid x, o_{i,<t})}{\pi_{\theta_{\text{old}}}(o_{i,t} \mid x,o_{i,<t})}$ is the importance sampling ratio.
The coefficient $\sigma(d_{j,t}) = \pi_\theta(d_{j,t}|x,d_{j,<t})\cdot\left(1-\pi_\theta(d_{j,t}|x,d_{j,<t})\right)$ stabilizes training with demonstration, and we set it to be $1$ for all on-policy steps.
We summarize our method as pseudocode in Alg.~\ref{alg:main}.

\section{Experiments}

\begin{table*}[htbp]
\centering
\begin{tabular}{@{}rccccc@{}}
\toprule
\multicolumn{1}{l}{}                            & Bamboogle        & MuSiQue          & Frames           & GAIA      & WebwalkerQA      \\ \midrule
\multicolumn{1}{l}{\textit{related works*}} &                  &                  &                  &                  &                  \\
GiGPO~\citeyearpar{feng2025group}& 68.90\%          & 18.90\%          &        -          &         -         &          -        \\
DeepResearcher~\citeyearpar{zheng2025deepresearcherscalingdeepresearch}& 72.80\%          & 29.30\%          &      -            &         -         &          -        \\ 
\midrule
\multicolumn{1}{l}{\textit{backbone}: Qwen3-8B} &                &                  &                  &                  &                  \\
ReAct Prompting                       &        50.93\%          &      18.64\%     &     31.92\%        &     9.69\%          &     18.62\%       \\
\multicolumn{1}{r}{Qwen3-8B-v3-SFT}             & 62.13\%          & 20.07\%          & 37.50\%          & 10.51\%          & \underline{19.85\%}          \\
+RFT                                             & 64.27\%          & 22.49\%          & 40.65\%          & \textbf{14.98\%}          & 19.56\%          \\
+DAPO/Search-R1                                  & \underline{70.83\%}          & \underline{23.60\%}          & \underline{43.59\%}          & 10.10\%          & 19.51\%          \\
\rowcolor[HTML]{e5f3fc} 
+ProCeedRL                                       & \textbf{73.87\%} & \textbf{29.52\%} & \textbf{46.42\%} & \underline{13.79\%} & \textbf{23.01\%} \\
\midrule
\multicolumn{1}{l}{\textit{ablation:}} Rewinding More    &  65.87\% & 19.11\% & 40.13\% & 10.10\% & 20.16\% \\
\bottomrule
\end{tabular}
\vspace{-0.5em}
\caption{Performance of different methods on deep search tasks.
\textbf{Bold} and \underline{underlined} results represent the best and the second best on each benchmark in our controlled experiments.
We report the average accuracy of three runs.
Our method outperforms baseline methods across most benchmarks, particularly on challenging ones such as MuSiQue.
\\ *For related works, we cite the metrics reported or run open-sourced models (if available) in our settings only for reference, as tool configurations like search engines differ across different works, which affect final performance.}
\label{tab:deepsearch_main}
\vspace{-10pt}
\end{table*}

In this section, we design extensive experiments to investigate:
\begin{enumerate}[itemsep=-2pt, topsep=6pt, leftmargin=1.5em]
    \item Does ProCeedRL improve the overall performance over baselines when variables are controlled (Sec.~\ref{sec:experiment_results})?
    \item What are the benefits of introducing ProCeedRL during test time and training (Sec.~\ref{sec:analysis})?
    \item How do the key design factors affect the performance of ProCeedRL(Sec.~\ref{sec:ablation_study})?
\end{enumerate}

\subsection{Experiment Settings}

\paragraph{Datasets and Benchmarks}
We assess ProCeedRL on two types of challenging agentic reasoning tasks: deep search QA and embodied agents.

In deep search QA, or search augmented QA, LLMs must automatically plan and invoke a search tool to find a final answer to the QA across multiple turns.
Building on HotpotQA~\citep{yang2018hotpotqa}, we first categorize the queries in its training set by topic.
Then, we obtain a curated training subset of 4000 question-answer pairs with balanced difficulties and topics as our training set.
We evaluate our model on several challenging question answering benchmarks: MuSiQue~\citep{trivedi2022musiquemultihopquestionssinglehop}, WebWalkerQA~\citep{wu2025webwalker}, GAIA~\citep{mialon2023gaia}, Frames~\citep{krishna2024factfetchreasonunified}, and Bamboogle~\citep{press2023measuringnarrowingcompositionalitygap}.

We also evaluate our method on the embodied benchmark ALFWorld~\citep{ALFWorld20}.
ALFWorld comprises a broad range of household settings in which LLMs must plan with a long horizon to solve different tasks.
It includes 3553 training settings, covering various household configurations and instructions, and we evaluate on both in- and out-of-distribution test set splits.
More details on the artifacts are provided in Appx.~\ref{appendix:artifacts}.

\paragraph{Implementation} Our implementation is based on verl~\citep{sheng2024hybridflow} and rllm~\citep{rllm2025}.
We adapt the ReAct~\citep{yao2023reactsynergizingreasoningacting} prompting framework.
For deep search QA, we formulate searching and answering as function calls.
We use the search engine You, and utilize the top 3 results' snippets as responses to each query.
To improve the model's multi-turn function-calling ability, we cold-start with samples generated by \texttt{DeepSeek-V3-0528}, which is utilized as the critic in our method.
We adopt the AlfWorld formulation as in \citet{feng2025group} and use the model itself as the critic.
We instruct both critics to output an integer rating between 0 and 10 in a zero-shot manner, with a threshold of 3.

We train our model using DAPO~\citep{yu2025dapoopensourcellmreinforcement} with a batch size of $32$ and a group size of $8$.
Half of each group is collected via ProCeedRL, with the other half collected directly.
For Qwen3-8B on AlfWorld, we additionally train our method with SFT on the correct samples collected by our method during rollout.
More details and hyperparameters are provided in Appx.~\ref{appendix:implementation_detail}.

\paragraph{Models and Baselines}
We conduct experiments on \texttt{Qwen3-1.7B/8B}~\citep{qwen3technicalreport}, and compare ProCeedRL with the following baselines:
\begin{enumerate}[itemsep=-3pt, topsep=5pt, leftmargin=1.5em]
    \item \textbf{SFT-only}: The model fine-tuned with samples collected from the critic's reasoning, to rule out the effect of knowledge distillation. 
    Additionally, we train the model using RFT~\citep{yuan2023scalingrelationshiplearningmathematical} on the correct data via rejection sampling from expert-generated trajectories; each group has the same size as in our method.
    \item \textbf{RL algorithms}: We compare with LLM-based agents trained with standard exploration RL, like GRPO and DAPO.
    In particular, we adjust the group size for each prompt to align the computational requirements for generation across methods.
    This is equivalent to several related works~\citep{song2025r1searcherincentivizingsearchcapability,feng2025retoolreinforcementlearningstrategic} that first successfully apply RLVR to agent domains.
    \item \textbf{Other baselines}: We also include several other related works~\citep{feng2025retoolreinforcementlearningstrategic,zheng2025deepresearcherscalingdeepresearch,dong2025arpo} on LLM agent reasoning for reference, as shown in Tab.~\ref{tab:deepsearch_main}.
    
\end{enumerate}

\subsection{Results}
\label{sec:experiment_results}

\begin{table}[htbp]
\centering
\resizebox{0.95\columnwidth}{!}{%
\begin{tabular}{@{}rcc@{}}
\toprule
\multicolumn{1}{l}{}                   & \multicolumn{2}{c}{AlfWorld}          \\ 
\multicolumn{1}{l}{}                   & in distribution & out of distribution \\ \midrule
\multicolumn{1}{c}{Qwen3-1.7B} & 11.07\%         & 12.69\%             \\
DAPO                                   & 18.69\%         &         \textbf{24.12\%}            \\
\rowcolor[HTML]{E5F3FC} 
ProCeedRL                              &      \textbf{20.35\%}      &     23.33\%          \\ \midrule
\multicolumn{1}{c}{Qwen3-8B}   & 44.22\%       & 47.07\%               \\
RFT                                    & 47.38\%      & 50.25\%                \\
DAPO                                   &    45.23\%          &     53.24\%              \\
\rowcolor[HTML]{E5F3FC} ProCeedRL                &            51.43\%     &       55.22\%              \\ 
\rowcolor[HTML]{E5F3FC} ProCeedSFT   &  \textbf{57.14\%}          &  \textbf{58.95\%}             \\
\bottomrule
\end{tabular}%
}
\caption{Main results on the embodied AlfWorld task.}
\label{tab:alfowrld}
\vspace{-1.5em}
\end{table}

\paragraph{Metrics} We employ the LLM-as-judge approach with \texttt{gpt-4o} to determine the correctness of generated answers for the deep search QA task, which we empirically find highly aligns with human judgment and gold reward.
For the embodied AlfWorld task, a solution's success is determined by the environment with a PDDL solver.
Details and prompts are provided in Appx.~\ref{appendix:prompt}.
We report the average accuracy of three runs.

\begin{figure*}[htbp]
    \centering
    \begin{minipage}[b]{0.59\textwidth}
        \includegraphics[width=\linewidth]{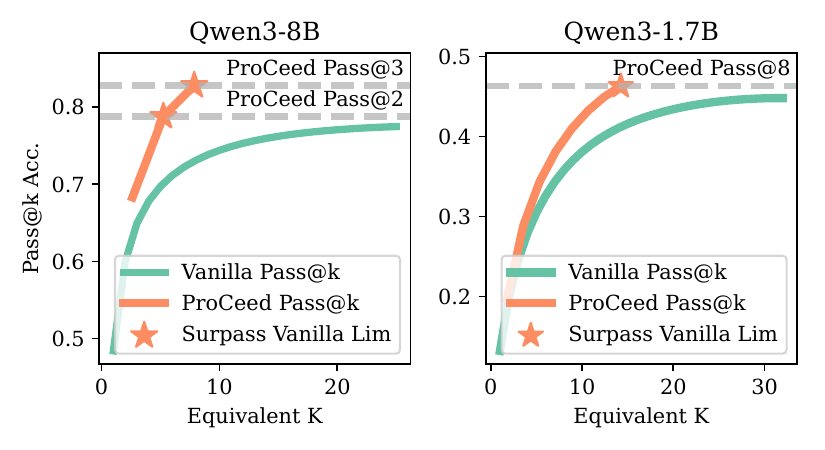}
        \vspace{-23pt}
        \caption{Pass$@k$ of ProCeed and vanilla rollout.
        The x-axis denotes the number of equivalent vanilla samples in terms of generation.
        Our method significantly improves exploration efficiency, matching pass$@k$ with less computation.
        Notably, it exceeds the saturation ceiling of vanilla exploration with a few samples (denoted by stars).%
        }
        \label{fig:exploration_upper_bound}
    \end{minipage}
    \hfill
    \begin{minipage}[b]{0.38\textwidth}
        \includegraphics[width=0.97\linewidth]{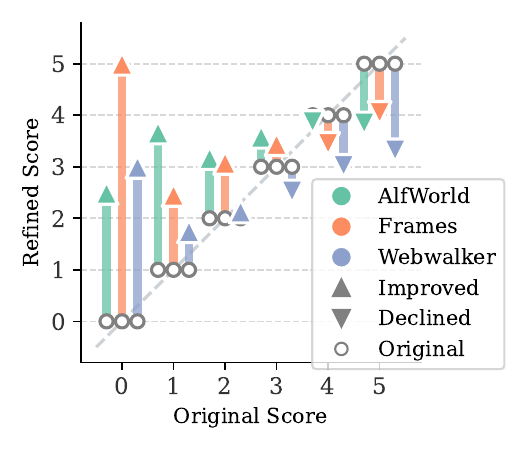}
        \vspace{-13pt}
        \caption{
        Improvement of refined actions over the original ones.
        Our method improves suboptimal actions with low ratings, whereas improvements gradually diminish as better original actions.
        }
        \label{fig:refined_demon}
    \end{minipage}
    \vspace{-10pt}
\end{figure*}

\paragraph{Main Results} 
The effectiveness of ProCeedRL is validated on deep search (Tab.~\ref{tab:deepsearch_main}) and embodied planning tasks (Tab.~\ref{tab:alfowrld}), where it substantially outperforms standard RL algorithms that utilize independent repeated sampling.
Specifically, ProCeedRL achieves an average improvement of 3.72\% on deep search QA, particularly on complex benchmarks such as MuSiQue, and increases performance by over 10\% on the embodied AlfWorld task when combined with SFT.
Furthermore, its superiority over RFT, which fine-tunes on expert trajectories selected via rejection sampling, underscores the critical role of on-policy data.
This result suggests that enhancing the model's reasoning capabilities is more effectively achieved by breaking the vicious cycle and improving exploration efficiency, rather than by merely distilling knowledge from complete, pre-validated expert solutions.

\subsection{Analysis and Ablation Study}

To further validate the importance of breaking the vicious circle between suboptimal actions and adverse contexts, and to demonstrate the effectiveness of our method, we conduct extensive analyses and ablation studies to support our design.

\subsubsection{Benefits of ProCeedRL}
\label{sec:analysis}

\paragraph{Exploration Effectiveness}
To verify that our method exceeds the model's reasoning upper bound due to the vicious circle, we test the pass$@k$ accuracy of ProCeed rollout against repeated sampling in AlfWorld using \texttt{Qwen3-1.7B/8B}.
The other settings are the same as in the main experiment.
We align the computational cost of pass$@k$ in Fig.~\ref{fig:exploration_upper_bound} to verify the exploratory gains of investing computation in this pipeline, rather than in generating additional trajectories.
Based on our statistics, with more details in Appx.~\ref{appendix:computation_overhead}, we find that one ProCeed trajectory costs approximately $2.5$ and $1.8$ vanilla samples for 8B and 1.7B models, respectively (e.g., ProCeed pass$@2$ is equivalent to vanilla pass$@5$ in terms of generations for the 8B model).

As shown in Fig.~\ref{fig:exploration_upper_bound}, ProCeed significantly improves exploration efficiency.
It matches vanilla pass$@k$ accuracy with substantially less computation, requiring only about half as many generations of samples for the 1.7B and 8B models.
Crucially, ProCeed surpasses the saturation ceiling of vanilla sampling with only a few samples ($2$ and $8$ samples for 8B and 1.7B models), demonstrating its ability to overcome the upper bounds of vanilla exploration.
This finding substantiates the benefits of our method and its capacity to break vicious cycles in exploration, thereby surpassing the upper bound of vanilla exploration.

\paragraph{Improvement of Refined Demonstration}
To directly quantify the efficacy of our refinement module, we evaluate the change in action quality for suboptimal steps identified by the critic.
We first isolate states with an initial critic score below five from the WebwalkerQA, Frames, and AlfWorld benchmarks.
For each state, we generate a refined action and then re-evaluate it using the same process critic, enabling a direct comparison of pre- and post-refinement scores.

As illustrated in Fig.~\ref{fig:refined_demon}, most steps with low initial scores show substantial improvement after refinement.
This enhancement is evident in the mitigation of error propagation from flawed intermediate actions to incorrect contextual information.
However, the utility of refinement displays a diminishing return as the quality of the original action increases.
Specifically, enforcing reflection on an already high-quality action can sometimes introduce confusion rather than refinement, potentially undermining performance.
These observations underscore the importance of determining an appropriate threshold for applying refinement, which we address in detail in a later discussion.

\begin{table}[htbp]
\centering
\resizebox{\columnwidth}{!}{
\begin{tabular}{@{}cccc@{}}
\toprule
Critic Type & Frames  & WebwalkerQA & \begin{tabular}[c]{@{}c@{}}AlfWorld\\ OOD\end{tabular} \\ \midrule
/           & 31.92\% & 18.64\%     & 47.07\%                                                \\
Self Critic & 38.83\% & 21.18\%     & 69.21\%                                                \\
Homogeneous & 37.54\% & 21.32\%     & 64.19\%                                                \\
External    & 41.18\% & 21.03\%     & 64.55\%                                                \\ \bottomrule
\end{tabular}}
\caption{ProCeed rollout accuracies with different critic choice.
All the critic settings improve test-time rollout accuracy, demonstrating the effectiveness and model-agnosticity of our method, as well as the importance of breaking the vicious circle during exploration.}
\label{tab:ablation_critic_choice}
\vspace{-1.5em}
\end{table}

\begin{figure*}[htbp]
    \centering
    \includegraphics[width=0.95\linewidth]{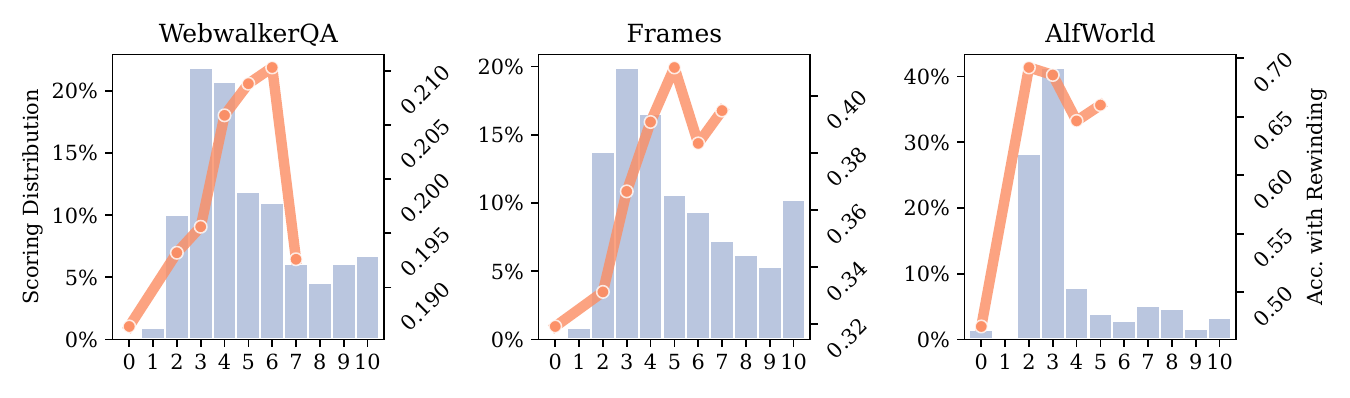}
    \vspace{-0.5em}
    \caption{
    Ablation on rewind thresholds ($l_{th}$).
    Blue histograms display the critic score distributions, and the orange line shows rollout accuracy for rewinding and refining all steps $\leq l_{th}$. While pruning low-scoring steps improves exploration, excessive rewinding at higher thresholds yields diminishing returns and performance degradation.
    }
    \label{fig:ablation_rewind}
    \vspace{-7pt}
\end{figure*}

\subsubsection{Ablation Study}
\label{sec:ablation_study}

\paragraph{Critic Choice}
The critic plays a vital role in the framework, identifying critical steps that significantly influence the exploration process.
To rule out the influence of the critic's knowledge, we test the following candidates:
\begin{enumerate}[itemsep=-4pt, topsep=4pt, leftmargin=1.5em]
    \item \textbf{Self-critic}, where the model critique itself on its performance;
    \item \textbf{Homogeneous Critic}, using a larger model of the same family;
    \item \textbf{External Critic}, where we leverage another powerful model.
\end{enumerate}
We accordingly employ \texttt{Qwen3-8B}, \texttt{Qwen3-30B\-A3B-Thinking-2507}, and \texttt{DeepSeek-V3-0528} as both critics and refinement policy.
ProCeed rollout accuracy for \texttt{Qwen3-8B} is evaluated across these settings on datasets from Section~\ref{sec:analysis}.

Tab.~\ref{tab:ablation_critic_choice} shows that all critic settings are effective, although with varying performance across benchmarks and models.
Crucially, the self-critic configuration alone yields substantial gains over the base model, even surpassing stronger models on specific tasks.
This highlights that the method’s efficacy relies on the procedural pipeline of identifying and correcting flawed actions to prevent cascading errors in reasoning contexts, rather than on oracle supervision.
In practice, we select critics comprehensively based on their exploration gains, as well as their similarity to the policy model, which helps ensure stable learning from demonstrations.

\paragraph{When to Rewind}
We conduct an ablation study of the rewind threshold, $l_\text{th}$, which determines the fraction of low-scoring trajectory steps to be refined.
To contextualize this analysis, we first examine the distribution of critic scores, which are assigned integer values from 0 to 10 for all steps, as shown in histograms in Fig.~\ref{fig:ablation_rewind}.
We measure the resulting ProCeed rollout accuracy by varying the threshold $l_\text{th}$ in our method, shown as the line chart in Fig.~\ref{fig:ablation_rewind}.
Other settings are held constant with the main experiments.

Results show that performance improves as more steps are reverted, but with diminishing returns.
This suggests that following the demonstration in a few steps is effective in preventing most cascading errors and misleading contexts.
More reverted steps help prune more incorrect steps; however, excessive reversion not only increases computational overhead but can also disrupt the model's reasoning by replacing adequate actions with suboptimal ones (as in Fig.~\ref{fig:refined_demon}).
We further validate this by training a model on deep search tasks with approximately twice as many reverted actions (as in Tab.~\ref{tab:deepsearch_main}).
Its significantly poorer performance corroborates our analysis.
These findings support our use of granular critic ratings over binary labels, as this approach enables fine-grained control over the process, and we set $l_\text{th}$ to balance all these effects in practice.

\section{Conclusions}

In this paper, we identify a ``vicious cycle'' in agentic reasoning in which suboptimal actions generate noisy feedback, degrade context, and hinder subsequent exploration.
To address this, we propose ProCeedRL, which employs a process critic to rewind and refine adverse steps in real time, thereby preventing error propagation.
Extensive experiments and analysis on deep search and embodied tasks demonstrate that ProCeedRL significantly outperforms standard RLVR baselines, surpassing its upper bounds in exploration and reasoning with superior efficiency.
Ultimately, ProCeedRL offers a scalable, self-contained method to advance agentic reasoning beyond the inherent upper bounds of independent repeated sampling.

\newpage

\section*{Limitations}
Our method incurs computational overhead compared with vanilla exploration RL.
The extra computation, introduced by the criticism and possible reflection at each step, incurs an asymptotic cost of $\mathcal{O}(1)$ per step and does not change the overall asymptotic behavior.
Based on our calculations, when the trajectory number is the same, the generation cost and time are roughly twice those of vanilla exploration (Appx.~\ref{appendix:computation_overhead}).
However, as we demonstrated, this improves exploration efficacy and exceeds the upper bound of vanilla exploration, thereby justifying the additional cost.
Moreover, our method does not require a critic at test time, although using one would improve reasoning.

Another limitation of our method is the lack of a guarantee of improvement.
Our method leverages the LLM's internal knowledge to identify suboptimal actions and refine them.
Although such an approach can improve step quality on average, it lacks theoretical guarantees and may lead to failures, thereby hindering exploration and affecting outcomes.
Applying hierarchical critics may help mitigate this issue, and we hope that future work on critics and LLM mechanisms can further address it and improve the scalability of our method.

Our method does not introduce additional risk.
While we use established public datasets that have mitigated privacy and toxicity risks, deploying autonomous agents may require ongoing safeguards to prevent unintended behavior in open-ended, real-world environments.

\bibliography{custom}

\appendix

\section{Algorithm}

We summarize our overall pipeline and algorithmic details as in Alg.~\ref{alg:main}.

\begin{algorithm}[ht]
\caption{ProCeedRL Training Loop}
\label{alg:main}
\begin{algorithmic}[1]
\Require Environment \texttt{env}, Policy $\pi_\theta$, Critic $\phi$, Refine Policy $\mu$, Threshold $l_\text{th}$, Max Steps $m_\text{max}$, Batch Size $bs$, Group Size $g$.

\For{Each Batch with $bs$}
    \State $\mathcal{B}\gets\{\}$
    \State $s\gets\ $\texttt{env.reset()}, $\tau\gets\{s_0\}$
    \For{Each Trajectory in Group $g$}
        \While{$i \leq m_\text{max}$}
            \State $a_{i}\gets \pi_\theta(\tau)$
            \If{Reversible \texttt{env}}
                \State $s_{i+1},\text{done}\gets\ $\texttt{env.step($a_i$)}
            \Else
                \State $s_{i+1}\gets None$
            \EndIf
            \State $l_t,c_t\gets\phi(\tau,a_i,s_{i+1})$
            \If{activate reflection and $l_t\leq l_\text{th}$}
                \State Rewind agent and environment
                \State $a_{i}\gets\mu(\tau,a_i,l_t,c_t)$
            \EndIf
            \State $s_{i+1},\text{done}\gets\ $\texttt{env.step($a_i$)}
            \State $\tau\gets\tau + \{a_i,s_{i+1}\}$, $i\gets i+1$
            \If {done}
                \State break
            \EndIf
        \EndWhile
        \State $\mathcal{B}\gets \mathcal{B}\cup \{\tau\}$
    \EndFor
    \State Calculate group-based $A_i$ as DAPO
    \State Update $\theta$ with $\mathcal{B}$ and Eq.~\ref{eq:optimization}.
\EndFor
\State \Return $\pi_\theta$
\end{algorithmic}
\end{algorithm}

\section{Computation for Critics}
\label{appendix:computation_overhead}

\begin{table}[htbp]
\centering
\resizebox{\columnwidth}{!}{%
\begin{tabular}{@{}cccccc@{}}
\toprule
\multirow{2}{*}{tasks}  & \multicolumn{2}{c}{policy}                           & \multicolumn{2}{c}{critic} & \multirow{2}{*}{ratio} \\
                        & model                     & tokens                   & type         & tokens      &                        \\ \midrule
\multirow{3}{*}{AlfWorld} & \multirow{2}{*}{Qwen3-8B} & \multirow{2}{*}{857.36} & thinking & 1320.59 & 2.54 \\
                        &                           &                          & instruct     & 211.69      & 1.25                   \\
                        & Qwen3-1.7B                & 1410.51                  & thinking     & 1111.39     & 1.79                   \\ \midrule
\multirow{2}{*}{Frames} & \multirow{2}{*}{Qwen3-8B} & \multirow{2}{*}{1010.36} & thinking     & 760.84      & 1.75                   \\
                        &                           &                          & instruct     & 246.02      & 1.24                   \\ \bottomrule
\end{tabular}%
}
\caption{The computation overhead per step for critics under different settings.}
\label{tab:computation_overhead}
\end{table}

The step-level critic in our method requires an additional generation stage to obtain feedback, resulting in an increased computational cost relative to vanilla exploration.
Here, we have tallied the average number of tokens generated by the agent and critic per step during actual operation to quantify computational overhead, as shown in Tab.~\ref{tab:computation_overhead}.
We calculate the ratio of computation required for collecting a trajectory in our method and in standard ways as $1 + \frac{\sharp(\text{average critic tokens})}{\sharp(\text{average policy tokens})}.$
We find that our method incurs computational overhead, with each sample requiring generations equivalent to $1.8$ to $2.5$ standard trajectories.
But such costs in each sample result in overall improvement in exploration efficiency and saturation performance, as discussed in Sec.~\ref{sec:analysis}.

We also collect the previous metrics when switching the critic to a non-thinking one, as shown in the ``instruct critic'' rows of Tab.~\ref{tab:computation_overhead}.
The reduced ratio suggests that we can also mitigate computational costs by replacing the critic with a non-thinking instruction model that produces shorter outputs.

\section{Implementation Details}
\label{appendix:implementation_detail}

We implement our training based on verl~\citep{sheng2024hybridflow} and rllm~\citep{rllm2025}, and we adopt the formulation of the AlfWorld benchmark from verl-agent~\citep{feng2025group}.
We use the default thinking mode for the \texttt{Qwen3-1.7B/8B} models.
We select the critic by comprehensively evaluating candidates based on their exploration gains and their similarity to the strategy model, thereby stabilizing learning with demonstrations.
The key parameters for running different algorithms and environment settings are listed in the following tables (Tab.~\ref{tab:sft_params},~\ref{tab:rl_params},~\ref{tab:env_params}).

\begin{table}[h!]
\centering
\begin{tabular}{@{}cc@{}}
\toprule
                        & value \\ \midrule
epochs                  & 3     \\
learning rate           & 1e-5  \\
batch size              & 64    \\
max tokens per step     & 4096   \\
temperature & 0.7   \\
top p      & 0.95  \\ \bottomrule
\end{tabular}
\caption{Key parameters for SFT}
\label{tab:sft_params}
\end{table}

\begin{table}[h!]
\centering
\begin{tabular}{@{}cc@{}}
\toprule
                    & value \\ \midrule
KL penalty          & 0.01  \\
train batch size    & 16   \\
ppo micro batch size & 256  \\
group size          & 8     \\
learning rate       & 1e-6  \\
temperature         & 0.7   \\
top p               & 1.00  \\
max length per step & 4096  \\ \bottomrule
\end{tabular}
\caption{Key parameters for RL}
\label{tab:rl_params}
\end{table}

\begin{table}[h!]
\centering
\begin{tabular}{@{}cc@{}}
\toprule
          & value (deep search/embodied) \\ \midrule
max steps & 10 / 50                      \\
threshold & 3                            \\ \bottomrule
\end{tabular}
\caption{Key parameters for environment configuration}
\label{tab:env_params}
\end{table}

Our experiments are conducted on Nvidia H100 and A100-80G GPUs.
The overall computation budget takes about $10$ gpu-months, most of which are conducted with $2$ for 1.7B models and $8$ for 8B models.

\section{Artifact Details}
\label{appendix:artifacts}

\paragraph{Bamboogle\citep{press2023measuringnarrowingcompositionalitygap}}
Bamboogle, distributed under an MIT License, contains 125 multi-hop questions to test an agent's ability to do multi-turn searching and knowledge combination.
The dataset is used in alignment with its license and the intended purpose of evaluating whether models can combine known facts to answer 2-hop questions.
The dataset is constructed through a human-verification process using popular Wikipedia entities, thereby mitigating potential privacy or harm issues.

\paragraph{MuSiQue\citep{trivedi2022musiquemultihopquestionssinglehop}}
MuSiQue is a large-scale English benchmark and is released under a CC BY 4.0 license.
It contains approximately 2,500 test questions, each requiring an answer based on 2 to 4 distinct paragraphs.
Its use is consistent to test an agent's ability to perform multi-turn, search-augmented reasoning.
The dataset was generated based on the Wikipedia corpus and does not contain private or toxic content.

\paragraph{GAIA\citep{mialon2023gaia}}
GAIA is a benchmark that aims to evaluate next-generation LLMs with augmented tool capabilities.
It includes 450 non-trivial questions with unambiguous answers, ranging in difficulty and requiring different levels of tooling and autonomy to solve.
We test on the 165 validation set.
It is publicly released on Hugging Face, and our use aligns with its intended purpose.

\paragraph{Frames\citep{krishna2024factfetchreasonunified}}
Frames is an English benchmark for retrieval agents, comprising 824 multi-hop questions that require handling complex constraints.
Released under the Apache-2.0 license, our use aligns with the license and its purpose of testing the factuality and reasoning ability of search agents.
The dataset was built using Wikipedia articles, ensuring high content quality.
It mitigates risks associated with private personal information or offensive material by selecting neutral, factual topics.

\paragraph{WebwalkerQA\citep{wu2025webwalker}}
WebwalkerQA is a bilingual (English and Chinese) benchmark comprising 680 search agent queries, released under the Apache-2.0 license.
It is designed and used to evaluate autonomous web navigation and long-context decision-making in dynamic environments.
The dataset was constructed by crawling publicly accessible webpages across safe, distinct domains to ensure that private user data or other sensitive personal information was excluded.

\paragraph{AlfWorld\citep{ALFWorld20}}
ALFWorld contains interactive TextWorld~\citep{cote2019textworldlearningenvironmenttextbased} environments that parallel embodied household environments in the ALFRED~\citep{shridhar2020alfredbenchmarkinterpretinggrounded} dataset.
AlfWorld tests an agent's ability to reason and learn high-level policies in an abstract space before solving embodied tasks through low-level actuation.
It contains 3553 training configurations, 120 in-distribution configurations, and 134 out-of-distribution configurations.
It is released under the MIT license and does not contain any personal information.

\paragraph{HotPotQA~\citep{yang2018hotpotqa}}
HotpotQA is a foundational dataset for diverse, explainable multi-hop question answering, released under a CC BY-SA 4.0 license, with 113k Wikipedia-based English question-answer pairs.
Its questions require finding and reasoning over multiple supporting documents, aligned with our purpose.
We used a subset of approximately 4000 questions, balanced for topic and difficulty, as our training set, thanks to its diverse question set.

\paragraph{Qwen3 Models~\citep{qwen3technicalreport}}
We utilize some models in the Qwen3 family in our experiment and training, including \texttt{Qwen3-1.7B}, \texttt{Qwen3-8B}, and \texttt{Qwen3-30B-A3B-Thinking-2507}.
These models are distributed under the Apache-2.0 license.
Our use in research aligns with their intended use.

\section{Prompts}
\label{appendix:prompt}

Given the environment settings and the agent's tasks across different environments, we design a separate prompt for the process-level critic.
It is based on our principles for evaluating steps by both their reasoning and their consequences in the following context, as discussed in Sec.~\ref{sec:method_framework}.
In practice, we ask the critic to produce additional analysis as CoT to improve rating accuracy.

\begin{tcolorbox}[title={Deep Search QA critic prompt}]
You are an expert in solving difficult problems through reasoning and retrieval.

Currently, an agent is attempting to answer a question through multi-round calls to search tools. The question, the agent's current action and retrieved documents, and history turns are provided.

Your task is to evaluate and score one round of the agent's calls, determining whether the search queries in the current round is correct for answering the question.

The score ranges from 0 to 10, where 0 means completely incorrect and 10 means completely correct.

Your score and analysis will serve as feedback to help the agent decide whether to revise the search terms or proceed to the next round of retrieval and reasoning.
If feeling necessary, you can also output your revised search terms and a supporting thinking process for these search queries.

Output format: The final output is list data in JSON format as 

\verb|`|\verb|`|\verb|`| json

\{\{"score": score, "critique": "your analysis", "suggestion\_search\_keywords": "[your revised queries,...]", "suggestion\_search\_reasoning": "step-by-step reasoning that supports the improved query as if you are the agent for it to learn from, without mentioning the current tool call and you are a critic."\}\}

\verb|`|\verb|`|\verb|`|

Question: \{problem\}

Search Query: \{tool\_results\}

History Turns: \{history\}
\end{tcolorbox}

\begin{tcolorbox}[title={AlfWorld critic prompt}]
You are an expert evaluator observing an agent trying to complete a task in a household environment.

You will be provided with the agent's task description, the environment's configuration, a history of its previous actions and observations, current admissible actions, and the agent's most recent action.

- Agent's Task: \{task\_description\}

- Environment Configuration: \{environment\_config\}

- History Turns (observations and the corresponding actions the agent took): \{action\_history\}

- Admissible Actions: \{admissible\_actions\}

- Agent's Latest Observation: \{latest\_observation\}

- Agent's Latest Action: \{agent\_action\}

Your goal is to evaluate and score the quality of the agent's latest action, determining if it is a good action for accomplishing the task. The score ranges from 0 to 10, where 0 means completely incorrect and 10 means completely correct.
Also give a consice yet complete analysis of the agent's last action. Was it logical? Did it move closer to the goal? Did it make a mistake?
If feeling neccessary (like the action is misleading), you can suggest a better candidate action FROM admissible actions together with your step-by-step reasoning. Your suggested action and reasoning may be used by the agent to improve its performance, so ensure the action is from the admissible actions and the reasoning is from an expert agent's first-person view.

Output Format: Output your critic result in JSON format as

\verb|`|\verb|`|\verb|`|json

{{"score": {{score}}, "critique": "{{your analysis}}", "suggestion\_action": "{{The better admissible action you suggest}}", "suggestion\_thought": "{{Your detailed step-by-step reasoning for the better action.}} "}}

\verb|`|\verb|`|\verb|`|

Ensure to enclose the entire JSON output within a single markdown code block.
\end{tcolorbox}

As for refinement prompts, we use a similar structure for both tasks, with the only difference being the instructions for action output, which is based on the task settings in different environments.

\begin{tcolorbox}[title=Refining Prompt 1]
An expert critic has commented on your latest step {latest\_step}, judging the quality of your latest step and whether it helps finishing your task.

- Critic's Overall Ratings (0-10): \{score\}.

- Critic's Analysis on your previous step: \{critic\_content\}

Your admissible actions of the current situation are: \{admissible\_actions\}.

Carefully read and understand the critic's feedback, and it's your turn to refine the step and retake an feasible action based on the feedback.

You should first reason step-by-step about the previous situation. This reasoning process MUST be enclosed within <think> </think> tags, and do not include any information about the critic, as if it's your first time making the action.

Once you've finished your reasoning, you should choose an admissible action for current step and present it within <action> </action> tags.
\end{tcolorbox}

\begin{tcolorbox}[title=Refine Prompt 2]
A critic has read and analysed your previous step:

- Critic's Overall Ratings (0-10): \{critique\_score\}

- Critic's Analysis on your previous step: \{critique\_content\}

- Critic's Suggestion search keywords: \{critique\_suggestion\}

Based on the critic's feedback, redo your previous tool call to improve its correctness and quality, in order to better solve the problem.

Note: When you redo and refine the tool call, you should act as if you are making the action the first time, do not add any information about the critic.
\end{tcolorbox}

\begin{tcolorbox}[title = {LLM-as-judge prompt}]
You are an impartial judge evaluating the correctness of an AI assistant's answer.

[Question]

\{question\}

[Correct Answer]

\{reference\_answer\}

[Assistant's Answer]

\{assistant\_answer\}

Task: Determine if the assistant's answer is correct by comparing it to the correct answer.

Instructions:

1. Extract the final answer from the assistant's response

2. Compare it with the correct answer

3. Provide your reasoning

4. Answer with "yes" if correct, "no" if incorrect
\end{tcolorbox}

\end{document}